\documentclass[letterpaper]{article}
\usepackage{aaai}
\usepackage{times}
\usepackage{amsmath} 
\usepackage{helvet}
\usepackage{courier}
\usepackage{url}
\usepackage{color}
\usepackage{graphicx}
\usepackage{amsfonts}
\usepackage{multirow}
\usepackage{booktabs}
\usepackage{arydshln}
\usepackage{mathabx}
\usepackage[normalem]{ulem}
\usepackage{soul}
\usepackage[subtle]{savetrees}
\usepackage[inline]{enumitem}
\usepackage{xspace}
\makeatletter
\newcommand*{\toll}[1]{$\overline{\hbox{#1}}\m@th$}
\makeatother
\let\oldbibliography\thebibliography
\renewcommand{\thebibliography}[1]{%
  \oldbibliography{#1}%
  \setlength{\itemsep}{0pt}%
}

\newcommand{\wA}{\widetilde{A}}
\newcommand{\wQ}{\widetilde{Q}}
\newcommand{\wC}{\widetilde{C}}

\newcommand{\daanet}{\textsc{D\kern-.15em A\kern-.15em A\kern-.15em Net}\xspace}
\newcommand{\bCA}{\overline{C}_{\mathrm{A}}}
\newcommand{\bCQ}{\overline{C}_{\mathrm{Q}}}
\newcommand{\shW}{\mathbf{W}_{\tiny{\mathrm{shared}}}}
\newcommand{\shb}{b_{\tiny{\mathrm{shared}}}}
\DeclareRobustCommand{\hlq}[1]{{\sethlcolor{yellow}\hl{#1}}}
\DeclareRobustCommand{\hla}[1]{{\sethlcolor{green}\hl{#1}}}
\begin{document}
%
\title{Dual Ask-Answer Network for Machine Reading Comprehension}
\author{
Han Xiao,\, Feng Wang,\, Jianfeng Yan,\, Jingyao Zheng\\
Tencent AI Lab\\
\{hanhxiao, madwang, larryjfyan, geozheng\}@tencent.com
}
\nocopyright
\maketitle
\begin{abstract}
\begin{quote}
There are three modalities in the reading comprehension setting: question, answer and context. The task of question answering or question generation aims to infer an answer or a question when given the counterpart based on context. We present a novel two-way neural sequence transduction model that connects three modalities, allowing it to learn two tasks simultaneously and mutually benefit one another. During training, the model receives question-context-answer triplets as input and captures the cross-modal interaction via a hierarchical attention process. Unlike previous joint learning paradigms that leverage the duality of question generation and question answering at data level, we solve such dual tasks at the architecture level by mirroring the network structure and partially sharing  components at different layers. This enables the knowledge to be transferred from one task to another, helping the model to find a general representation for each modality. The evaluation on four public datasets shows that our dual-learning model outperforms the mono-learning counterpart as well as the state-of-the-art joint models on both question answering and question generation tasks.
\end{quote}
\end{abstract}

\section{Introduction}
Recently the task of machine reading comprehension~\cite{rajpurkar2016squad,nguyen2016ms} has received increasing attention from NLP and AI research communities. The task attempts to enable machines to answer questions after reading a passage. There are three modalities in this task: question, answer and context. Many successful question answering models have been proposed to fill answer modality by modeling the interactions between questions and contexts modalities~\cite{seo2016bidirectional,yu2018qanet}. Intuitively, question and answer are quite similar in the sense that they are both short texts and strongly related to the given context. Few very recent works have recognized such relationship and explored from different perspectives~\cite{song2017unified,wang2017joint,tang2018learning}. Their promising results suggest that:
\begin{enumerate*}[label=(\roman*)]
\item the roles of questions and answers are switchable given context, in the sense that the same network structure can be used for both question answering (QA) and question generation (QG); 
\item this invertibility or duality can be exploited to improve QA and QG.
\end{enumerate*}

\begin{figure}[htb!]
\centering
\includegraphics[width=1.0\linewidth]{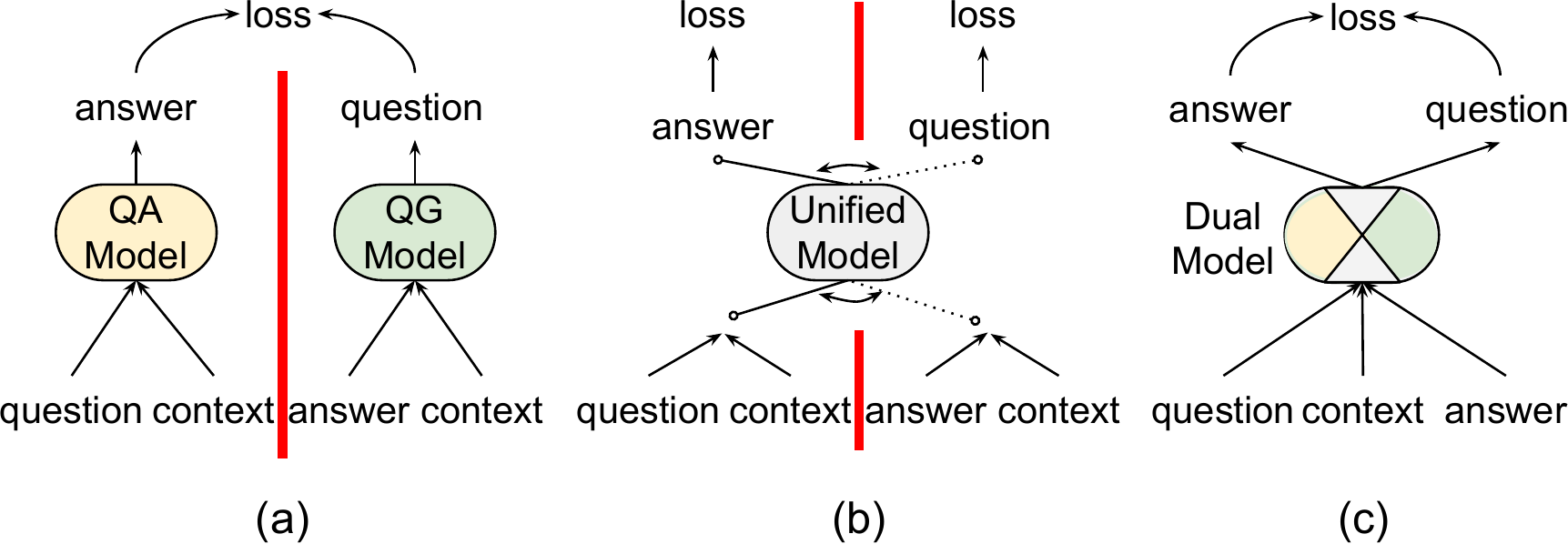}
\caption{Three learning paradigms to exploit the task correlations of question answering and question generation. Red line represents data/model-level separation. (a) Two separated models with a joint loss function~\cite{song2017unified}. (b) A unified model with alternated training input and two separated loss functions~\cite{wang2017joint}. (c) \textbf{This work}: a unified architecture with locally shared structure that can learn two tasks simultaneously. In contrast to (a) and (b), there is no data-level or model-level separation in our learning paradigm.}
\label{fig:duality}
\end{figure}


In this work, we consider machine reading comprehension as a dual learning problem, i.e. learning to answer versus learning to ask. It is no surprise that to utilize the commonality among the tasks, some sharing scheme is required in the learning paradigm. Figure~\ref{fig:duality} visualizes previous dual learning paradigms and the one from this work. Specifically, we propose a novel two-way neural sequence transduction model that jointly solves these two tasks. The model is structurally symmetric and its components are shared between two tasks at different levels, as illustrated in Figure~\ref{fig:overview}. During training, the model receives question-context-answer triplets as input and captures the cross-modal interaction via a hierarchical attention process. During testing, the model generates answer or question as sequence when given the counterpart based on context. Our contributions are summarized as follows:
\begin{figure*}[htb!]
\centering
\includegraphics[width=.9\textwidth]{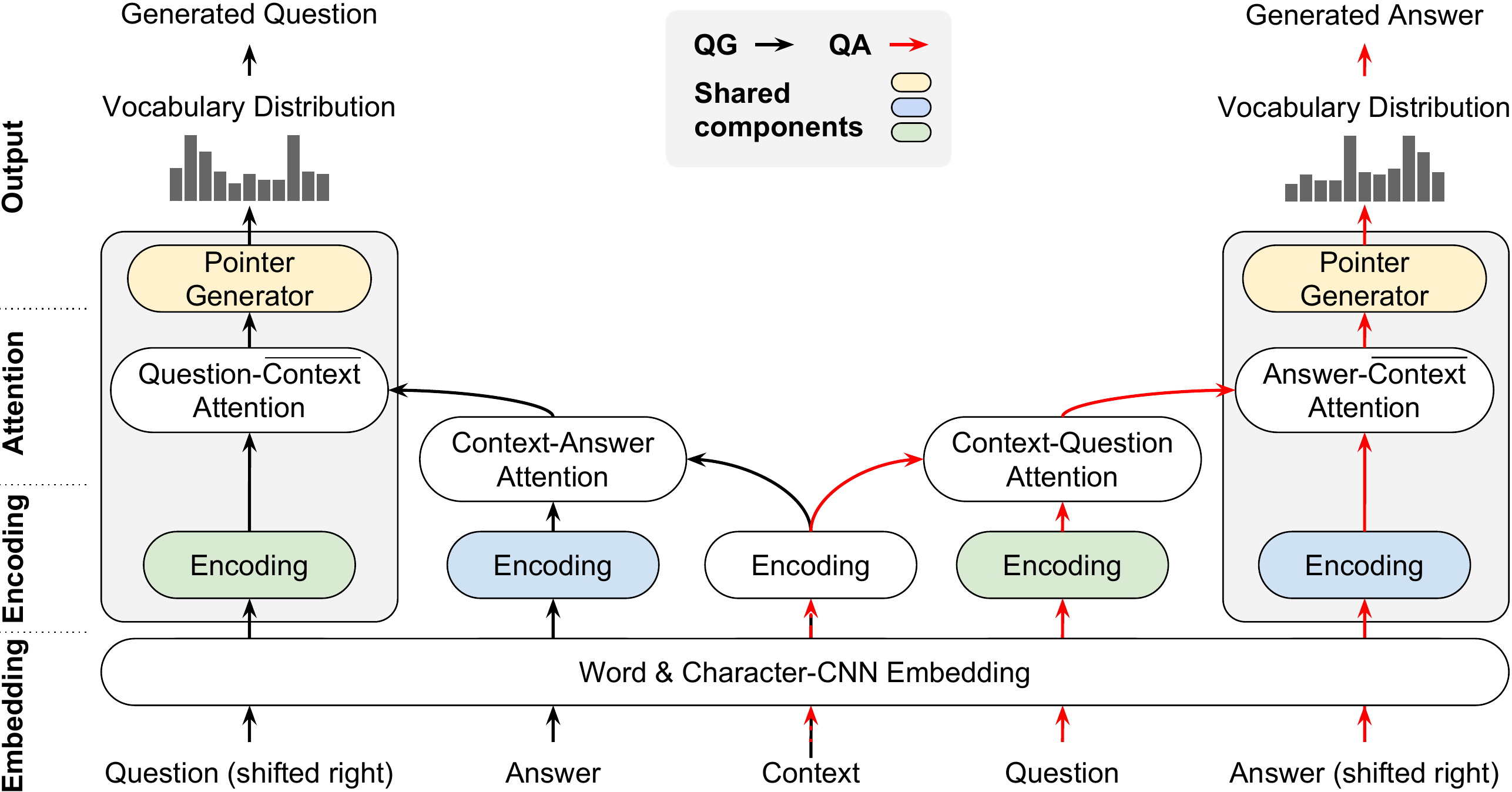}
\caption{The model architecture of Dual Ask-Answer Network (\daanet). \emph{Best viewed in color}. It is a hierarchical process consisted of four layers: embedding, encoding, attention and output. The rectangle super-block on the side can be viewed as a decoder for QG and QA, respectively. The computational flow of QG and QA task is plotted with black arrows and red arrows, respectively. Shared (including partially shared) components are filled with the same color, i.e. blue for the answer encoder, green for the question encoder and yellow for the pointer generator. Note that, the context encoder is also shared by QA and QG. For the sake of clarity we only draw one context encoder block. During testing, the shifted input is replaced by the model's own generated words from the previous steps.}
\label{fig:overview}
\end{figure*}
\begin{itemize}
\item We propose a novel two-way neural sequence transduction model that learns question answering and question generation simultaneously. We tie the network components that playing similar roles in the two tasks to transfer cross-task knowledge during training. The cross-modal interaction of question, context and answer is captured by a pair of symmetric hierarchical attention processes. To the best of our knowledge, we are the first to leverage duality at this granularity under the machine reading comprehension setting.
\item We demonstrate the effectiveness of our model on four public datasets and achieve promising results. Our model outperforms the mono-learning counterpart and the state-of-the-art joint question answering models. We provide a strong evidence for bringing the duality into model structures to improve the reading comprehension ability. The code and data used in this paper are available online to enable future comparison\footnote{\url{https://github.com/hanxiao/daanet}}.
\end{itemize}

\section{Related Work}
Our work relates to the existing works on machine reading comprehension (MRC), question answering (QA) and question generation (QG). In this section, we briefly review the previous works from these perspectives.

A great number of MRC models have been developed over the past few years, e.g. BiDAF~\cite{seo2016bidirectional}, S-Net~\cite{tan2017s}, R-Net~\cite{wang2017gated}, match-LSTM~\cite{wang2016machine}, ReasonNet~\cite{shen2017reasonet}, Document Reader~\cite{chen2017reading}, Reinforced Mnemonic Reader~\cite{hu2017reinforced}, FusionNet~\cite{huang2017fusionnet} and QANet~\cite{yu2018qanet}. Most of the existing models rely on the assumption that the answer is a continuous span of a given passage. Under this assumption, an answer can be simplified as a pair of two integers, representing its start and end position in the passage respectively. In this work, we do not make such assumption. Our model generates both answer and question as sequence, so that the same model architecture can be used for both question answering and question generation. In general, we believe such generative model offers better expressivity for complicated semantics, and thus is more likely to deliver a true breakthrough in machine reading comprehension.


A large part of prior works on question generation relies on feature engineering, handcrafted templates and linguistic rules~\cite{aldabe2006arikiturri,heilman2011automatic,liu2010automatic,bordes2014open,dhingra2018simple}. 
Recently, encouraged by the remarkable success of the sequence-to-sequence model in machine translation~\cite{sutskever2014sequence,luong2015effective}, parsing~\cite{vinyals2015grammar} and text summarization~\cite{nallapati2016abstractive}, the interest of using deep neural network to tackle QG rises rapidly~\cite{du2017learning,yuan2017machine,duan2017question}. One of the major motivations in these studies is using QG to enrich the training data of QA. For example, a corpus with 30M QA pairs is generated by transducing facts from a knowledge base into natural language questions using a deep neural network~\cite{serban2016generating}.


Finally, some very recent MRC works have recognized the relationship between QA and QG and exploited it differently. In spite of a similar objective shared by this study, our work is unique from the following perspectives: 
\begin{itemize}
\item Unlike~\cite{yang2017semi} that rely on the continuous span assumption of answer and~\cite{sachan2018self,tang2017question,tang2018learning} that consider answering as a sentence selection task, we consider both question answering and question generation as sequence transduction so that it is possible to use the same model architecture for both tasks. Unlike the extraction-then-synthesis framework proposed in~\cite{tan2017s}, our model is trained completely end-to-end.
\item Unlike~\cite{song2017unified} that train a QA model and a QG model independently using the same architecture and~\cite{wang2017joint} that alternate the training data between QA and QG examples for the same model, our two-way model directly consumes question-context-answer triplets during training. Given a triplet, the parameters will be updated ``twice'' in the sense that they receive gradients from both QA and QG directions. Consequently, the parameters in our model are trained more sufficiently comparing to the other two related works.
\item Unlike~\cite{tang2018learning} that use a collaboration detector to connect the training of two tasks, we propose a novel sharing scheme to leverage duality at the architecture level. The components and parameters that playing similar roles in the two tasks are tied together. The interaction between question, answer and context is captured by coupling two symmetric hierarchical attention processes. This sharing scheme greatly reduces the total number of parameters required for the two tasks.
\end{itemize}

\section{Dual Ask-Answer Network}
We first formulate the dual learning problem of asking and answering in the MRC setting, and then present our model: Dual Ask-Answer Network (\daanet). Finally, we make a summary about the attention and the duality used in the model.

\subsection{Problem Formulation}
Unlike the traditional MRC problem focusing only on QA, the problem considered in this work is bipartite: QA and QG. Specifically, the model should be able to infer answers or questions when given the counterpart based on context. 

Formally, we denote a context paragraph as $C:=\{c_1, \ldots, c_n\}$, a question sentence as $Q:=\{q_1,\ldots,q_m\}$ and an answer sentence as $A:=\{a_1,\ldots,a_k\}$. In the sequel, we follow this notation and use $n$, $m$, $k$ to represent the length of a context, a question and an answer, respectively. The context $C$ is shared by the two tasks. Given $C$, the QA task is defined as finding the answer $A$ based on the question $Q$; the QG task is defined as finding the question $Q$ based on the answer $A$. In contrast to previous MRC models that mostly assume the answer as a continuous span, we regard both QA and QG tasks as generation problems and solve them jointly in a neural sequence transduction model.

\subsection{Model Description}

The high level architecture of our proposed Dual Ask-Answer Network is illustrated in Figure~\ref{fig:overview}. This neural sequence transduction model receives string sequences as input and processes them through an embedding layer, an encoding layer, an attention layer, and finally to an output layer to generate sequences.

\subsubsection{1. Embedding Layer.} The embedding layer maps each word to a high-dimensional vector space. The vector representation includes the word-level and the character-level information. The parameters of this layer are shared by context, question and answer. For the word embedding, we use pre-trained $256$-dimensional GloVe~\cite{pennington2014glove} word vectors, which are fixed during training. All the out-of-word vocabulary words are mapped to an \texttt{<UNK>} token. Besides that, there are three special tokens: \texttt{<PAD>}, \texttt{<START>} and \texttt{<END>}. The embeddings of \texttt{<START>}, \texttt{<END>} and \texttt{<UNK>} are trainable and initialized randomly, whereas \texttt{<PAD>} is fixed to an all-zero vector. 

For the character embedding, each character is represented as a $200$-dimensional trainable vector. Consequently, each word can be represented as a sequence of character vectors, where the sequence length is either truncated or padded to $16$. Next, we conduct 1D CNN with kernel width $3$ followed by max-pooling along the time axis. This gives us a fixed-size $200$-dimensional vector for each word. Finally, the concatenation of the character embedding and word embedding vectors is linearly projected to a 300-dimensional space and then passed to a highway network~\cite{srivastava2015highway} as follows:
\begin{align*} 
&\mathbf{e}:=[\mathbf{e}_\mathrm{word}, \mathbf{e}_\mathrm{char}]\mathbf{H}_1 + v_1,\quad \mathbf{g}:=\sigma(\mathbf{e}\mathbf{H}_2 + v_2),\\
&\mathbf{e}_\mathrm{out}:=\mathbf{g}\odot\mathbf{e} + (1-\mathbf{g})\odot(\mathbf{e}\mathbf{H}_3 + v_3),
\end{align*}
where $\{\mathbf{H}_1\in \mathbb{R}^{456\times300}$, $\mathbf{H}_2,\mathbf{H}_3\in \mathbb{R}^{300\times300}$ and $v_1,v_2,v_3\in \mathbb{R}\}$ are learned parameters; $\sigma$ denotes the sigmoid function. The dimension of the output is $300$. 

\subsubsection{2. Encoding Layer.}
The encoding layer contains three encoders for context, question and answer, respectively. They are shared by QA and QG, as depicted in Figure~\ref{fig:overview}. That is, given QA and QG dual tasks, the encoder of the primal task and the decoder of the dual task are forced to be the same. The parameter sharing scheme serves as a regularization to influence the training on both tasks. It also helps the model to find a more general and stable representation for each modality.

\begin{figure}[htb!]
\centering
\includegraphics[width=.95\linewidth]{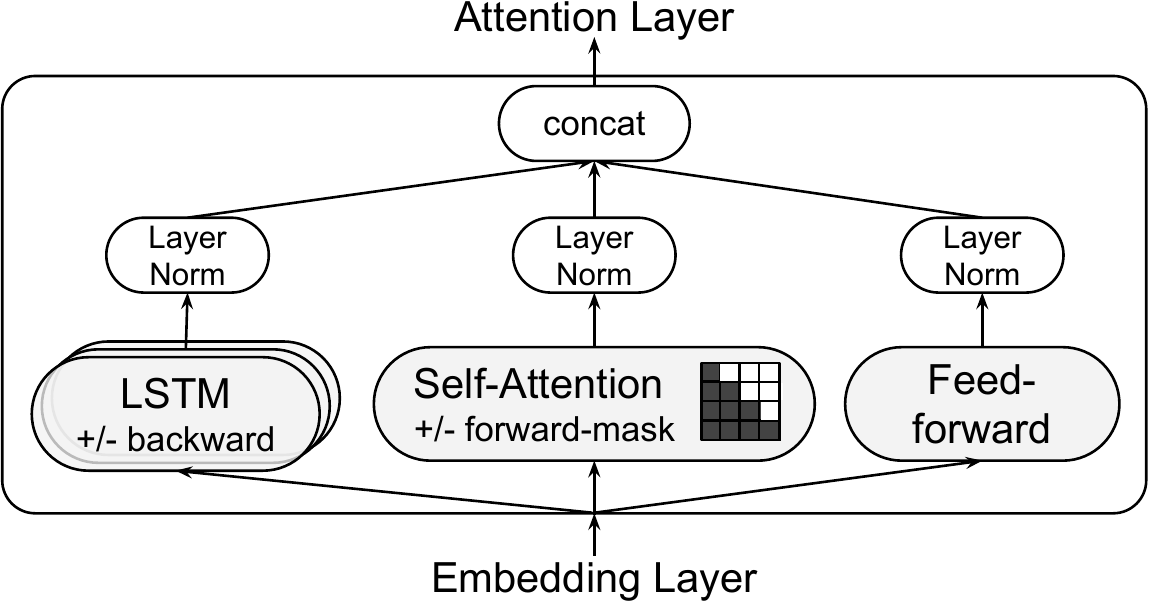}
\caption{The architecture of a context/question/answer encoder. In the question and answer encoder, LSTM is unidirectional and a forward mask is used when computing self-attention.}
\label{fig:encode}
\end{figure}

Each encoder consists of the following basic building blocks: a element-wise fully connected feed-forward block, stacked LSTMs~\cite{gers1999learning} and a self-attention block~\cite{vaswani2017attention}. The self-attention block allows each position in the encoder to attend to all positions based on the similarity measured by their dot products. The attention function used in self-attention block is described in the attention layer. Each block is followed by layer normalization~\cite{ba2016layer}. The final output of an encoder is a concatenation of the outputs of all blocks, as illustrated in Figure~\ref{fig:encode}.

The context encoder consists of a feed-forward network, a 3-layer bidirectional LSTM and a 4-head self-attention. The question and answer encoders follow the same composition but are with necessary modifications to prevent the ``leakage'' of leftward information. Specifically, in question and answer encoder LSTM is \emph{unidirectional} and a forward mask is used when computing self-attention. The forward mask keeps only attention of later position to early position, meanwhile setting the remaining to $-\infty$ as follows,
\begin{align*}
\mathrm{SelfAtt}(\mathbf{e}_\mathrm{out})&:=\mathrm{softmax}\big(\frac{1}{\sqrt{d}}\mathbf{e}_\mathrm{out}\mathbf{R}(\mathbf{e}_\mathrm{out}\mathbf{R})^\top + \mathbf{M}\big)\mathbf{e}_\mathrm{out},\\
\mathbf{M}_{ij}&:=0\quad\mathrm{if}\ i<j\quad\mathrm{else}\  -\infty,
\end{align*}
where $\mathbf{e}_\mathrm{out}$ is the output of the embedding layer and $\mathbf{R}\in \mathbb{R}^{300\times d}$ is a learned parameter. Different compositions of the encoder are benchmarked in the experiment section. 

\subsubsection{3. Attention Layer.} 
\begin{figure}[htb!]
\centering
\includegraphics[width=.95\linewidth]{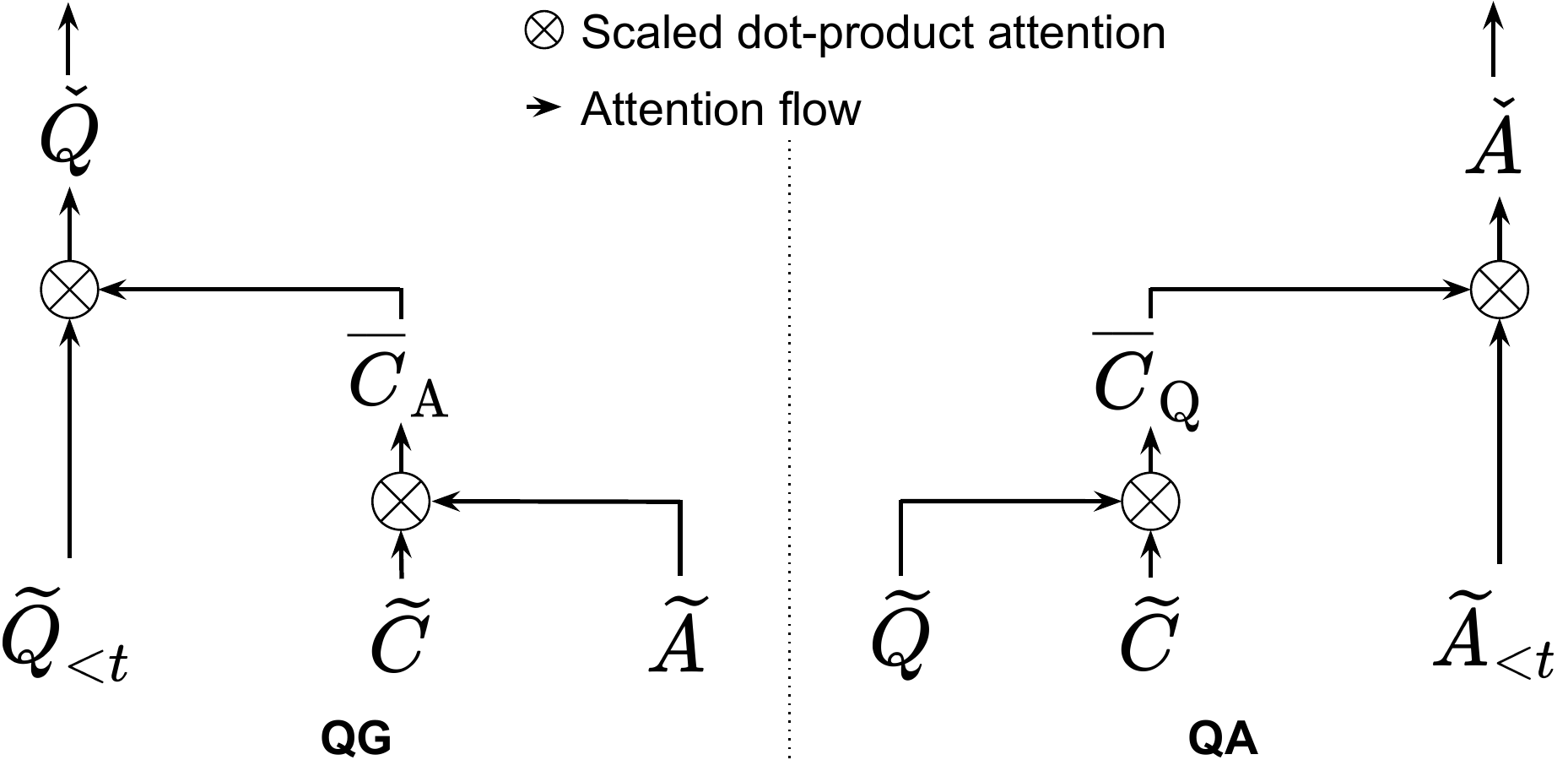}
\caption{The two-step attention flow implemented in the attention layer, where the input $\wQ,\wC,\wA,\wQ_{<t},\wA_{<t}$ are from the previous encoding layer.}
\label{fig:attention}
\end{figure}

The attention layer fuses all information observed so far from context encode $\wC$, answer encode $\wA$ and question encode $\wQ$ in a hierarchical manner. At time $t$ in QG, we first fold the answer encode into the context encode, which is folded again into the question encode $\wQ_{<t}$ generated previously. The QA part follows the similar attention procedure. In this work, we denote the first fold-in step as the ``context-$\ast$'' attention and the second fold-in step as ``$\ast$-\toll{context}'' attention, where \toll{context} is the output of the first fold-in step. Figure~\ref{fig:attention} visualizes this two-step attention procedure.

More concretely, the context-answer attention is computed as follows: we first compute the normalized similarities between each pair of context and answer word as $s(\wC,\wA):=\mathrm{softmax}(\frac{1}{\sqrt{d}}\wC \mathbf{U}(\wA \mathbf{V})^\top)$, where the score function $s:\mathbb{R}^{n\times d}\times \mathbb{R}^{k\times d}\mapsto \mathbb{R}^{n\times k}$. Learnable variables in $s$ are $\mathbf{U},\mathbf{V}\in\mathbb{R}^{d\times d^{\prime}}$. This is the multiplicative attention with the scaling factor of $\frac{1}{\sqrt{d}}$, as proposed in~\cite{vaswani2017attention}. The context-answer attention is then computed as a weighted sum of the answer, i.e. $\bCA:=s(\wC,\wA)\cdot \wA$. The context-question attention follows the same procedure, i.e. $\bCQ:=s(\wC,\wQ)\cdot \wQ$. The output of the first fold-in step is in  $\mathbb{R}^{n\times d}$, the same size as the context sequence. 

The second fold-in step involves computing question-\toll{context} and answer-\toll{context} attention, which is essential for generating meaningful question and answer sequence, respectively. It allows every position in the generated sequence to attend over all positions in \toll{context} sequence from the first fold-in step. At time $t$, the input of question-\toll{context} attention consists of $\wQ_{<t}$ from the previous encoding layer and $\bCA$ from the context-answer attention, the output is $\widecheck{Q}:=s(\wQ_{<t},\bCA)\cdot \bCA$. The answer-context attention follows the same procedure, i.e. $\widecheck{A}:=s(\wA_{<t},\bCQ)\cdot \bCQ$.

\subsubsection{4. Output Layer.} The output layer generates an output sequence one word at a time. At each step the model is auto-regressive~\cite{graves2013generating}, consuming the words previously generated as input when generating the next. In this work, we employ the pointer generator as the core component of the output layer~\cite{see2017get}. It allows both copying words from the context via pointing, and sampling words from a fixed vocabulary. This aids accurate reproduction of information especially in QA, while retaining the ability to generate novel words.

\begin{figure}[htb!]
\centering
\includegraphics[width=.99\linewidth]{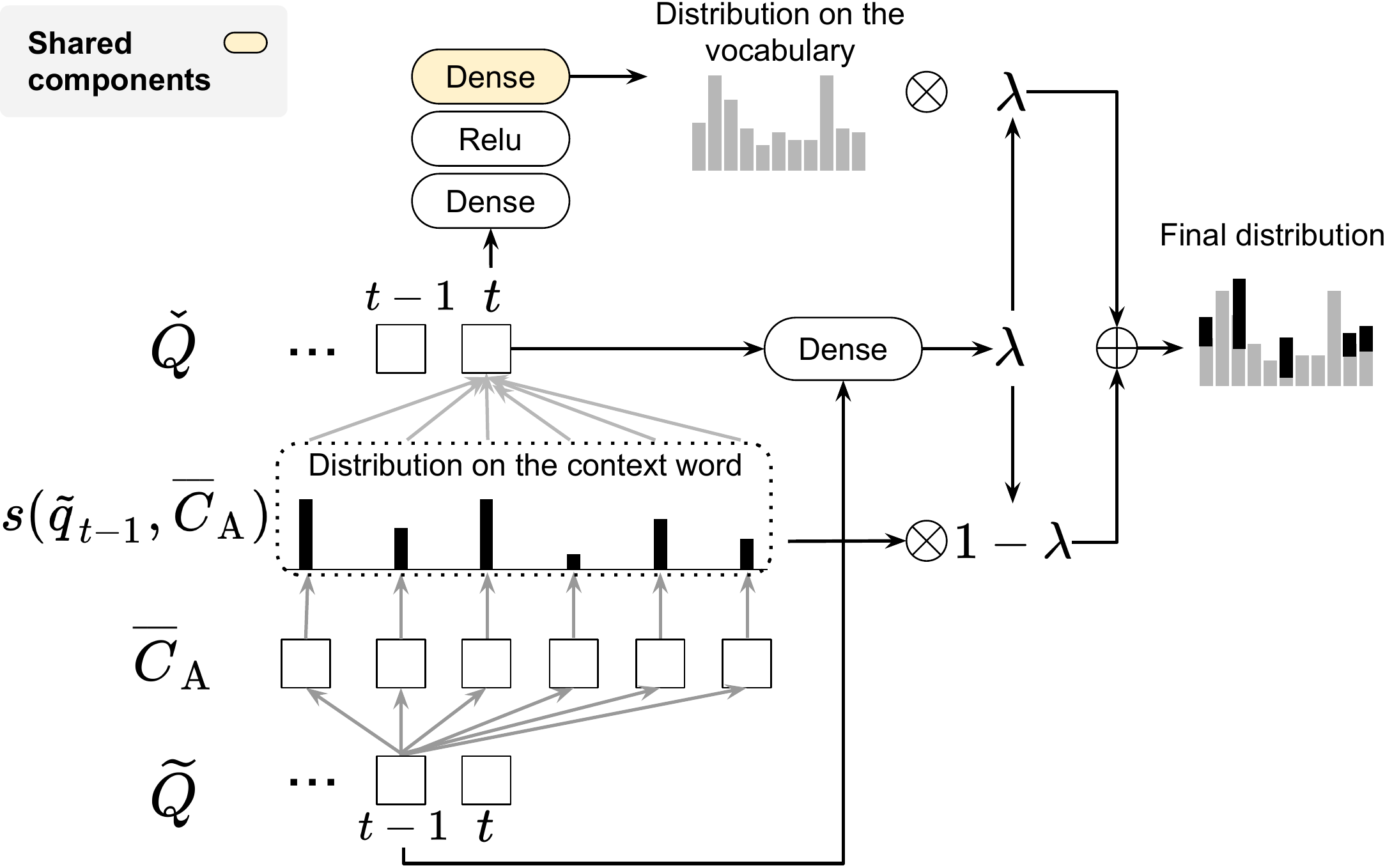}
\caption{The architecture of the output layer for QG. Each square block represents a word. The final distribution is a mixture of two discrete distributions. QA follows the same architecture with few replacements. The dense layer in yellow is shared by QA and QG.}
\label{fig:output}
\end{figure}

The architecture of the output layer is depicted in Figure~\ref{fig:output}. Specifically, at time $t$ the probability of generated word $w$ is defined as a mixture of two discrete distributions $P_t(w):=\lambda P^{\mathrm{vocab}}_{t}(w)+ (1-\lambda)P^{\mathrm{context}}_{t}(w)$, where $P_t^{\mathrm{vocab}}(w)$ is a distribution over the predefined vocabulary, $P_t^{\mathrm{context}}(w)$ is a distribution over words in the current context $C$ based on the ``$\ast$-\toll{context}'' attention score (duplicated words are merged), and $\lambda_t$ is a soft switch to choose between generating a word and copying from the context. Formally, the final vocabulary distribution for QG follows the form below:
\begin{align*} 
P^{\mathrm{vocab}}_{t}&:=\mathrm{softmax}\big(\tanh( \widecheck{q}_{t}\mathbf{W}_{1}+b_1)\shW+\shb\big)\\
P^{\mathrm{context}}_{t}&:= s(\widetilde{q}_{t-1},\bCA)\\
\lambda_{t}&:=\sigma([\ \widecheck{q}_{t},\ \widetilde{q}_{t-1}]\mathbf{W}_2 + b_2),
\end{align*}
where $s(\cdot, \cdot)$ is the score function defined in the attention layer; $\widecheck{q}_{t}$ and $\widetilde{q}_{t-1}$ corresponds to the $t^\mathrm{th}$ row of $\widecheck{Q}$ and the $(t-1)^\mathrm{th}$ row of $\widetilde{Q}$, respectively. Learnable variables are $\mathbf{W}_{1}\in \mathbb{R}^{d\times1024}$, $\shW\in \mathbb{R}^{1024\times|V|}$, $\mathbf{W}_{2}\in \mathbb{R}^{2d\times1}$, $b_1\in \mathbb{R}^{1024}$, $\shb\in \mathbb{R}^{|V|}$ and a scalar $b_2$, where $|V|$ represents the number of words in the vocabulary. Note that $\shW$ and $\shb$ are shared by QA and QG, whereas $\{\mathbf{W}_1,\mathbf{W}_2,b_1,b_2\}$ are task-specific parameters.

There are two advantages of this sharing scheme. First, sharing $\shW$ significantly reduces the number of parameters, which helps the model to better capture the cross-modal interaction in the attention layer. Second, it forces the model to first project the decoded information from their own spaces to a common latent space before projecting to the vocabulary space. This common latent space serves as a tie that connects two tasks, providing another channel for transferring knowledge. One may also consider it as a regularization we set at the upper-level of the network to influence the training process. In the experiment section, we show that this sharing scheme indeed contributes and improves the performance.

For QA, the parametric form of $P_t(w)$ is defined similarly by first replacing $\widetilde{q}_{t-1}$ with $\widetilde{a}_{t-1}$, $\bCA$ with $\bCQ$, and $\widecheck{q}_{t}$ with $\widecheck{a}_{t}$; then equipping the task-specific parameters $\{\mathbf{W}_1,\mathbf{W}_2,b_1,b_2\}$; finally sharing $\shW$ and $\shb$ with QG.

\subsection{Loss Function}
During training, the model is fed with a question-context-answer triplet $(Q, C, A)$, and the decoded $\widehat{Q}$ and $\widehat{A}$ from the output layer are trained to be similar to $Q$ and $A$, respectively. To achieve that, our loss function consists of two parts: the negative log-likelihood loss widely used in the sequence transduction model and a coverage loss to penalize repetition of the generated text, which is akin to the form in~\cite{see2017get}. 

We employ \emph{teacher forcing} strategy in training, where the model receives the groundtruth token from time $(t-1)$ as input and predicts the token at time $t$. Specifically, given a triplet $(Q, C, A)$ the objective is to minimize the following loss function with respect to all model parameters:
\begin{equation*}
\begin{split}
\ell :=& \underbrace{-\sum_{t=1}^{m}\Big(\log p(\widehat{q}_t=q_t\,|\, Q_{<t}, A, C) - \kappa \overbrace{\sum_{\min}(s_{t}^{\mathrm{Q\overline{C}}}, \sum_{t'=1}^{t-1}s_{t'}^{\mathrm{Q\overline{C}}})}^{\mbox{coverage loss of QG}}\Big)}_{\mbox{QG loss}}\\
&\underbrace{-\sum_{t=1}^{k}\Big(\log p(\widehat{a}_t=a_t\,|\, A_{<t}, Q, C) - \kappa \overbrace{\sum_{\min}(s_t^{\mathrm{A\overline{C}}}, \sum_{t'=1}^{t-1}s_{t'=1}^{\mathrm{A\overline{C}}})}^{\mbox{coverage loss of QA}}\Big)}_{\mbox{QA loss}},
\end{split}
\end{equation*}
where $s_t^{\mathrm{Q\overline{C}}}$ and $s_t^{\mathrm{A\overline{C}}}$ corresponds to the $t^\mathrm{th}$ row of question-\toll{context} and the answer-\toll{context} attention score obtained from the second fold-in step in the attention layer, respectively; $\sum_{\min}$ represents the sum of element-wise minimum of two vectors; $\kappa$ is a hyperparameter for weighting the coverage loss. 

\subsection{Attention in the Model}
Our model exploits the attention mechanism in three places.

\begin{itemize}
  \item In the encoding layer, we employ self-attention as one way to capture long-range dependencies. Comparing to LSTM, the maximum path length a signal has to traverse in the network is much shorter in self-attention, making it easier to learn long-range dependencies. When encoding context, we allow both forward and backward signals to traverse in the self-attention network. When encoding question and answer with self-attention, we prevent the backward signal with a mask to preserve the auto-regressive property.
  \item In the attention layer, we develop a two-step attention that folds in all information observed so far for generating final sequences. The first fold-in step captures the interaction between  question/answer and context and represents it as a new \toll{context} sequence. The second fold-in step mimics the typical encoder-decoder attention mechanisms in the sequence-to-sequence models~\cite{bahdanau2014neural,gehring2017convolutional}.
  \item Finally in the output layer, we recycle the attention score obtained from the second fold-in step and use it as the copy distribution in the pointer generator. The ultimate vocabulary is thus defined as an interpolation of a distribution over the current context words and a distribution over the predefined large vocabulary. This facilitates copying words from the context, while retaining the ability to generate new words.
\end{itemize}

\subsection{Duality in the Model}
Our model exploits the duality of QA and QG in two places.
\begin{itemize}
\item As we consider both QA and QG are sequence generation problems, our architecture is reflectional symmetric, as depicted in Figure~\ref{fig:overview}. The left QG part is a mirror of the right QA part with identical structures. Such symmetry can be also found in the attention calculation and the loss function. Consequently, the answer modality and the question modality are connected in a two-way process through the context modality, allowing the model to infer answers or questions given the counterpart based on context.
\item Our model contains shared components between QA and QG at different levels. Starting from the bottom, the embedding layer and the context encoder are always shared between two tasks. Moreover, the answer encoder in QG is reused in QA for generating the answer sequence, and vice versa. On top of that, in the the pointer generator, QA and QG share the same latent space before the final projection to the vocabulary space. The cycle consistency between question and answer is utilized to regularize the training process at different levels, helping the model to find a more general representation for each modality.
\end{itemize}

\section{Experimental Results}
\subsection{Implementation Details}
As every component in the proposed model is differentiable, all parameters could be trained via back propagation. We randomly initialized the parameters using the ``fan-avg'' strategy, i.e. sampling from a uniform distribution whose width is the average number of inputs and output connections~\cite{glorot2010understanding}.  We filtered out words that appear less than 5 times in the data and replaced them with \texttt{<UNK>} token, resulting a vocabulary with $90,000$ words. We fixed the word embedding to a pretrained $256$-dimensional GloVe word vectors~\cite{pennington2014glove}. The character embedding was in the size of $66\times256$ and was learned during training. Dropout was mainly applied to the encoding layer with the keep rate of $0.9$. The coverage loss weight $\kappa$ was $1.0$. We used the Adam optimizer~\cite{kingma2014adam} for minimizing the loss function. The gradient was clipped by restricting its $\ell_2$-norm less than or equal to $5.0$. The learning rate was increased from zero to $0.001$ with an inverse exponential function and then fixed for the remainder of training. The batch size was empirically set to $16$. During testing, we conduct auto-regressive decoding separately for QA and QG. The input at time $t$ is its own generated words from the previous steps $<t$. Decoding is terminated when the model encounters the first \texttt{<END>} or when the sequence contains more than $100$ words.

\subsection{Datasets}
Our experiments were conducted on four datasets: \texttt{SQuAD}~\cite{rajpurkar2016squad}, \texttt{MSMARCO}~\cite{nguyen2016ms}, \texttt{WikiQA}~\cite{yang2015wikiqa} and \texttt{TriviaQA}~\cite{JoshiTriviaQA2017}. We subsampling \texttt{MSMARCO} to speedup the experiments. All data are preprocessed to \texttt{SQuAD} format, which is available online. Table~\ref{tbl:data-summary} summarizes the statistics of the training and testing data used in the experiments.

\begin{table}[htb!]
\centering
\begin{tabular}{lrrrrrr}
\toprule
Dataset           & \#Train & \#Test & $\overline{n}$ & $\overline{m}$ & $\overline{k}$ \\ \midrule
\texttt{SQuAD}    & $86,821$& $5,928$& $117.1$& $10.1$& $3.1$\\
{\texttt{MSMARCO}}  &$120,000$& $24,000$& $405.8$& $4.1$& $12.3$\\
{\texttt{TriviaQA}} &$120,000$& $24,000$& $631.9$& $12.2$& $1.5$\\
\texttt{WikiQA}   &$873$& $126$& $471.2$& $6.4$& $24.1$\\
\bottomrule
\end{tabular}
\caption{Summary of the datasets. Statistics include the number of training/testing $(Q,A,C)$ triplets; the average number of words in context ($\overline{n}$), question ($\overline{m}$) and answer ($\overline{k}$).}
\label{tbl:data-summary}
\end{table}

\subsection{Evaluation Metrics}
As both question and answer are generated in our model, we adopt BLEU-1,2,3,4~\cite{papineni2002bleu} and Meteor~\cite{denkowski:lavie:meteor-wmt:2014} scores from machine translation, and ROUGE-L from text summarization~\cite{lin2004rouge} to evaluate the quality of the generation. These metrics are calculated by aligning machine generated text with one or more human generated references based on exact, stem, synonym, and paraphrase matches between words. Higher score is preferable as it suggests better alignments with the groundtruth.

\subsection{Effectiveness of Dual Learning}
Table~\ref{tbl:dual-vs-mono} summarizes the performance of our model versus the mono-learning counterpart (\texttt{mono}), JointQA~\cite{wang2017joint} (\texttt{jqa}), multi-perspective QG without reinforcement learning~\cite{song2017unified} (\texttt{mpqg}) and a simple sequence-to-sequence model with attention mechanism (\texttt{s2s}). Both \texttt{jqa} and \texttt{mpqg} share the same objective as our work, i.e. leveraging the duality of QA and QG to improve each other. We implement \texttt{mono} by masking out all the task-irrelevant structures and parameters from \daanet. One can observe that, \daanet considerably outperforms the mono-learning counterpart with an average margin of $5.7$pp in Rouge-L and $5.3$pp in Bleu-4 on QA. It also outperforms the state-of-the-art joint model in most of the cases. This result provides a strong evidence for bringing the duality into model structures to improve the reading comprehension ability. One may also notice that, the improvement of \daanet is more significant in QA than in QG. This is due to the fact that there are many ways to convey the same question in QG, e.g. by replacing interrogative, thus metrics based on literally matching may underestimate the quality of generated question. We leave the investigation of a better QG performance metric as a future work.

\begin{table}[htb!]
\setlength{\tabcolsep}{1.5pt}
\centering
\fontsize{7.5}{7.2}
\begin{tabular}{llrrrrr|rrrrr}
\toprule
 & & \multicolumn{5}{c|}{QA} & \multicolumn{5}{|c}{QG} \\
 & & \tiny{\daanet} & \small{\texttt{mono}} & \small{\texttt{jqa}} & \scriptsize{\texttt{mpqg}} & \small{\texttt{s2s}} 
   & \tiny{\daanet} & \small{\texttt{mono}} & \small{\texttt{jqa}} & \scriptsize{\texttt{mpqg}}&\small{\texttt{s2s}}\\
\midrule
\parbox[t]{2mm}{\multirow{6}{*}{\rotatebox[origin=t]{90}{\texttt{SQUAD}}}}
& \scriptsize{B1} & $\mathbf{37.28}$ & $29.62$ & $26.66$ & $14.65$ & $15.47$ & 
                        $33.95$ & $31.28$ & $\mathbf{34.48}$ & $29.92$ & $28.73$\\
& \scriptsize{B2} & $\mathbf{32.66}$ & $25.37$ & $22.14$ & $11.17$ & $11.44$ & 
                        $\mathbf{18.99}$ & $16.23$ & $18.84$ & $14.82$ & $13.05$\\
& \scriptsize{B3} & $\mathbf{29.31}$ & $22.23$ & $18.82$ & $8.74$  & $8.74$ & 
                        $\mathbf{12.35}$ & $9.92$  & $12.09$ & $8.64$ & $6.73$\\
& \scriptsize{B4} & $\mathbf{26.38}$ & $19.56$ & $16.11$ & $6.93$  & $6.75$ & 
                        $\mathbf{8.71}$  & $6.64$  & $8.32$ & $5.49$ & $3.74$\\
& \scriptsize{RL} & $\mathbf{43.85}$ & $37.38$ & $41.43$ & $26.96$ & $24.21$ & 
                        $\mathbf{35.58}$ & $33.32$ & $34.31$ & $32.24$ & $30.92$\\
& \scriptsize{Mt} & $\mathbf{23.21}$ & $19.73$ & $21.70$ & $12.64$ & $8.95$ & 
                        $\mathbf{14.18}$ & $12.52$ & $13.89$ & $11.70$ & $11.41$\\
\midrule
\parbox[t]{2mm}{\multirow{6}{*}{\rotatebox[origin=t]{90}{\texttt{MSMACRO}}}}
& \scriptsize{B1} & $\mathbf{44.98}$ & $41.20$ & $41.26$ & $35.51$ & $29.41$ & 
                        $58.45$ & $\mathbf{59.04}$ & $56.28$ & $49.42$ & $53.20$ \\
& \scriptsize{B2} & $\mathbf{38.09}$ & $34.70$ & $34.53$ & $28.51$ & $22.84$ & 
                        $45.65$ & $\mathbf{46.14}$ & $43.64$ & $37.25$ & $40.01$\\
& \scriptsize{B3} & $\mathbf{34.83}$ & $31.62$ & $31.37$ & $25.43$ & $19.66$ & 
                        $35.98$ & $\mathbf{37.47}$ & $34.14$ & $28.40$ & $30.31$\\
& \scriptsize{B4} & $\mathbf{32.82}$ & $29.70$ & $29.42$ & $23.60$ & $17.67$ & 
                        $28.79$ & $\mathbf{29.18}$ & $26.99$ & $22.10$ & $22.86$\\
& \scriptsize{RL} & $\mathbf{44.25}$ & $42.23$ & $41.30$ & $35.23$ & $32.75$ & 
                        $57.90$ & $\mathbf{58.19}$ & $56.13$ & $50.90$ & $53.57$\\
& \scriptsize{Mt}& $\mathbf{22.60}$ & $22.28$ & $22.06$ & $16.93$ & $15.33$ & 
                        $29.52$ & $\mathbf{29.78}$ & $27.94$ & $24.14$ & $25.73$\\
\midrule
\parbox[t]{2mm}{\multirow{6}{*}{\rotatebox[origin=t]{90}{\texttt{TriviaQA}}}}
& \scriptsize{B1} & $\mathbf{62.13}$ & $48.47$ & $47.67$ & $41.41$ & $50.14$ & 
                        $\mathbf{45.61}$ & $43.26$ & $37.57$ & $34.07$ & $31.93$\\
& \scriptsize{B2} & $\mathbf{59.40}$ & $45.13$ & $44.54$ & $37.91$ & $45.07$ & 
                        $\mathbf{33.25}$ & $29.91$ & $23.49$ & $19.91$ & $17.43$\\
& \scriptsize{B3} & $\mathbf{53.41}$ & $39.68$ & $40.90$ & $31.45$ & $32.88$ & 
                        $\mathbf{26.60}$ & $22.81$ & $16.91$ & $13.23$ & $10.45$\\
& \scriptsize{B4} & $\mathbf{40.93}$ & $29.10$ & $34.54$ & $20.10$ & $19.18$ & 
                        $\mathbf{22.32}$ & $18.33$ & $11.48$ & $9.46$ & $6.72$\\
& \scriptsize{RL} & $\mathbf{61.97}$ & $47.89$ & $46.86$ & $41.94$ & $51.87$ & 
                        $\mathbf{45.31}$ & $42.69$ & $38.26$ & $34.57$ & $32.81$ \\
& \scriptsize{Mt}& $\mathbf{38.80}$ & $29.64$  & $28.45$ & $24.79$ & $31.38$ & 
                        $\mathbf{21.35}$ & $19.67$ & $16.64$ & $14.54$ & $13.43$\\
\midrule
\parbox[t]{2mm}{\multirow{6}{*}{\rotatebox[origin=t]{90}{\texttt{WikiQA}}}}
& \scriptsize{B1} & $\mathbf{37.31}$ & $36.47$ & $31.74$ & $33.53$ & $7.49$ & 
                        $14.13$ & $13.49$ & $14.45$ & $\mathbf{15.53}$ & $11.06$ \\
& \scriptsize{B2} & $\mathbf{31.76}$ & $30.89$ & $26.31$ & $28.81$ & $2.82$ & 
                        $7.67$ & $7.28$ & $7.76$ & $\mathbf{7.78}$ & $6.37$\\
& \scriptsize{B3} & $\mathbf{29.74}$ & $28.78$ & $24.33$ & $27.02$ & $1.21$ & 
                        $2.89$ & $0.00$ & $0.00$ & $0.00$ & $0.00$\\
& \scriptsize{B4} & $\mathbf{28.61}$ & $27.56$ & $23.25$ & $25.98$ & $0.56$& 
                        $1.71$ & $0.00$ & $0.00$ & $0.00$ & $0.00$\\
& \scriptsize{RL} & $\mathbf{37.32}$ & $35.50$ & $29.91$ & $36.41$ & $10.74$& 
                        $18.86$ & $19.61$ & $19.24$ & $\mathbf{20.03}$ &$18.45$\\
& \scriptsize{Mt}& $19.50$ & $18.90$ & $14.95$ & $\mathbf{20.08}$ & $3.47$ & 
                        $\mathbf{5.71}$ & $5.71$ & $5.63$ & $5.47$ & $4.27$\\
\bottomrule
\end{tabular}
\caption{Results of question answering and question generation on four datasets. B$\ast$ represents Bleu-1,2,3,4; RL is Rouge-L; Mt is Meteor. Higher is better; best in bold.}
\label{tbl:dual-vs-mono}
\end{table}

\subsection{Ablation Study}
Table~\ref{tbl:ablations} summarizes the performance of \daanet and its ablations on our \texttt{SQuAD} test set, in which we compare different compositions of encoder, with and without attention mechanisms and different sharing schemes. First, one can observe that LSTM is the most important piece in encoding, without which the performance decreases around $10$pp on QA and $2.5$pp on QG. Self-attention also contributes to the improvement with $3$pp on QA and $0.6$pp on QG. Next, removing context-$\ast$ attention leads to a catastrophic performance. This is in line with our intuition as no context-$\ast$ attention means that the answer generation is independent to the given question and depends only on the given context. Finally, the last part of Table~\ref{tbl:ablations} demonstrates the effectiveness of the proposed sharing scheme.

\begin{table*}[htb!]
\centering
\begin{tabular}{lrrrr|rrrr}
\toprule
\multirow{ 2}{*}{Ablation} & \multicolumn{4}{c}{QA} & \multicolumn{4}{c}{QG} \\
   & Bleu-1 & Bleu-4 & Rouge-L & Metero& Bleu-1 & Bleu-4 & Rouge-L & Metero \\ \midrule
Encoder without LSTM  & $20.80$ & $11.22$ & $30.84$ & $14.84$ & $31.60$ & $6.30$ & $32.75$ & $12.38$\\
Encoder without self-attention & $33.24$ & $22.69$ & $39.90$ & $21.31$ & $33.56$ & $8.14$ & $34.10$ & $13.65$\\
No context-$\ast$ attention & $4.96$ & $0.38$ & $4.61$ & $1.81$ & $25.74$ & $2.05$ & $26.99$ & $8.47$\\
No copy mechanism & $8.24$ & $0.94$ & $11.23$ & $3.99$ & $26.32$ & $2.22$ & $27.29$ & $8.83$\\
Unshared question \& answer encoders & $29.82$ & $19.43$ & $37.75$ & $19.47$ & $33.23$ & $8.51$ & $35.26$ & $13.86$\\
Unshared context encoder & $22.90$ & $13.45$ & $29.33$ & $14.67$ & $33.37$ & $8.17$ & $34.94$ & $13.75$\\
Unshared output vocabulary projection & $32.64$ & $21.46$ & $40.37$ & $20.77$ & $33.71$ & $8.15$ &$34.77$ & $13.85$\\
\midrule
\daanet& $\mathbf{37.28}$ & $\mathbf{26.38}$ & $\mathbf{43.85}$ & $\mathbf{23.21}$ & $\mathbf{33.95}$ & $\mathbf{8.71}$ & $\mathbf{35.58}$ & $\mathbf{14.18}$\\
\bottomrule
\end{tabular}
\caption{The performance of \daanet and its ablations on \texttt{SQuAD}. Higher is better; best in bold}
\label{tbl:ablations}
\end{table*}

\subsection{Case Study}
Table~\ref{tbl:casestudy} presents some generated questions and answers from \daanet and the mono-learning models. Question or answer is generated given the gold counterpart based on context. In the first two samples, \daanet works perfectly, whereas the mono-learning model fail to provide the desired output. In the third sample, the generated question from \daanet is more readable comparing to the groundtruth. Empirically, we find that \daanet poses questions that are semantically similar to the referenced questions but phrased differently. Although human may consider these as good cases, \daanet still scores poorly on QG under all alignment-based evaluation metrics. This explains the scores on QG are overall lower than QA and the improvements on QG are less significant than on QA. We attach more generated samples in the supplementary document.

\begin{table}[htb!]
\centering
\begin{tabular}{p{.08\columnwidth}p{.83\columnwidth}}
\toprule
\small{Context} & \small{In the course of the \hla{10th century}, the initially destructive incursions of Norse war bands into the rivers of France evolved into more permanent encampments that included local women and personal property. \hlq{The Duchy of Normandy}, which \hlq{began in} \hla{911} as a fiefdom, was \hlq{established} by the treaty of Saint-Clair-sur-Epte between King...}\\
\small{Question} & {\tiny{$\triangleright$}} \small When was the Duchy of Normandy founded?\\
\small{Answer} & {\tiny{$\diamond$}} \small 911\\\hdashline[1pt/1pt]
\scriptsize{\daanet} & {\tiny{$\triangleright$}} \small when did the duchy of normandy open?\\
 & {\tiny{$\diamond$}} \small 911\\
\small\texttt{mono} & {\tiny{$\triangleright$}} \small when did duchy of normandy begin?\\
 & {\tiny{$\diamond$}} \small 10th century\\
\midrule
\small{Context} & \small{... Melbourne is home to a number of museums, art galleries and theatres and is also described as the "sporting capital of Australia". \hla{The Melbourne Cricket Ground} is \hlq{the largest stadium in Australia}, and the host of the 1956 Summer Olympics and the 2006 Commonwealth Games...}\\
\small{Question} & {\tiny{$\triangleright$}} \small What is the largest stadium in Australia?\\
\small{Answer} & {\tiny{$\diamond$}} \small Melbourne Cricket Ground\\\hdashline[1pt/1pt]
\scriptsize{\daanet} & {\tiny{$\triangleright$}} \small what is largest stadium in australia ?\\
 & {\tiny{$\diamond$}} \small melbourne cricket ground\\
\small\texttt{mono} & {\tiny{$\triangleright$}} \small what is largest gross state product (AFL)?\\
 & {\tiny{$\diamond$}} \small the melbourne cricket ground\\
\midrule
\small{Context} & \small{\hlq{Sky UK Limited} (\hlq{formerly} \hla{British Sky Broadcasting or BSkyB}) is a British telecommunications company which serves the United Kingdom. Sky provides television and broadband internet services and fixed line telephone services to consumers and businesses in the United Kingdom. It is the UK's \hlq{largest pay-TV} broadcaster with 11 million customers as of 2015...}\\
\small{Question} & {\tiny{$\triangleright$}} \small Sky UK Limited is formerly known by what name?\\
\small{Answer} & {\tiny{$\diamond$}} \small British Sky Broadcasting\\\hdashline[1pt/1pt]
\scriptsize{\daanet} & {\tiny{$\triangleright$}} \small what is name of sky uk limited?\\
 & {\tiny{$\diamond$}} \small british sky broadcasting or BSkyB\\
\small\texttt{mono} & {\tiny{$\triangleright$}} \small what is canada's largest pay-TV service?\\
 & {\tiny{$\diamond$}} \small BSkyB\\
\bottomrule
\end{tabular}
\caption{Selected outputs from \daanet and \texttt{mono}. Yellow text is question-related; green text is answer-related.}
\label{tbl:casestudy}
\end{table}

\section{Conclusion}
We proposed Dual Ask-Answer Network, a two-way neural sequence transduction model, which solves question answering and question generation for machine reading comprehension. We exploited the duality of QA and QG tasks by sharing local structures at different levels. The attention mechanism is used in multiple layers to capture the interaction of context, question and answer. We demonstrated the effectiveness of our model on four public datasets, providing a strong evidence for bringing the duality into model structures to improve the reading comprehension ability. One interesting future direction is to study the collaboration of multiple \daanet. For example, one can stack \daanet multiple times, such that the generated question and answer from the lower network are fed to the upper network. This ``bootstrapping'' strategy can be considered as an implicit way of data augmentation, alleviating the shortage on labeled MRC data.

\bibliography{cites}
\bibliographystyle{aaai}
\end{document}